\documentclass[
  reprint,
  amsmath,amssymb,
  aps,
  prapplied,
  superscriptaddress,
  longbibliography
]{revtex4-2}

\usepackage{graphicx}    
\usepackage{dcolumn}     
\usepackage{bm}          
\usepackage{hyperref}    
\usepackage{xcolor}
\usepackage{amsfonts}
\usepackage{booktabs}    
\usepackage{caption}
\usepackage{subcaption}
\usepackage{multirow}
\usepackage{tikz}
\usepackage{amsmath} 
\usetikzlibrary{calc, positioning, fit, arrows.meta, backgrounds}
\usepackage[utf8]{inputenc}
\usepackage{hyperref}
\usetikzlibrary{decorations.pathreplacing}

\pgfdeclarelayer{background}
\pgfdeclarelayer{foreground}
\pgfsetlayers{background,main,foreground}

\begin{document}

\title{Reservoir Computing as a Language Model}

\author{Felix Köster and Atsushi Uchida\\
\textit{Department of Information and Computer Sciences, Saitama University,}\\
\textit{255 Shimo-Okubo, Sakura-ku, Saitama City, Saitama, 338-8570, Japan}
}
\email{felixk@mail.saitama-u.ac.jp, auchida@mail.saitama-u.ac.jp}

\date{\today}

\begin{abstract}
Large Language Models (LLM) have dominated the science and media landscape duo to their impressive performance on processing large chunks of data and produce human-like levels of text.
Nevertheless, their huge energy demand and slow processing are still a bottleneck to further increasing quality while also making the models accessible to everyone.
To solve this bottleneck, we will investigate how reservoir computing performs on natural text processing, which could enable fast and energy efficient hardware implementations.
Studies investigating the use of reservoir computing as a language model remain sparse.
In this paper, we compare three distinct approaches for character-level language modeling, two different \emph{reservoir computing} approaches, where only an output layer is trainable, and the well-known \emph{transformer}-based architectures, which fully learn an attention-based sequence representation. We explore the performance, computational cost and prediction accuracy for both paradigms by equally varying the number of trainable parameters for all models. Using a consistent pipeline for all three approaches, we demonstrate that transformers excel in prediction quality, whereas reservoir computers remain highly efficient reducing the training and inference speed. Furthermore, we investigate two types of reservoir computing: a \emph{traditional reservoir} with a static linear readout, and an \emph{attention-enhanced reservoir} that dynamically adapts its output weights via an attention mechanism. Our findings underline how these paradigms scale and offer guidelines to balance resource constraints with performance.
\end{abstract}

\maketitle

\section{Introduction}
Modern sequence modeling tasks, such as language modeling and machine translation, have been dominated by attention-based architectures—most prominently the transformer family \cite{vaswani2017attention, devlin2019bert, radford2018improving, brown2020language, schick2020small, gao2020making}—which excel at capturing long-range dependencies by learning contextual representations through self-attention layers and feed-forward networks. However, transformer training and inference incur substantial computational and energy costs, often requiring specialized hardware and limiting accessibility under tighter resource budgets.

\begin{figure*}[t]
\centering
\begin{tikzpicture}[node distance=1cm, auto, >=Stealth]

  \node (letter1) [draw, rectangle, inner sep=5pt] {s};
  \node (letter2) [draw, rectangle, inner sep=5pt, right=of letter1] {h};
  \node (letter3) [draw, rectangle, inner sep=5pt, right=of letter2] {a};
  \node (letter4) [draw, rectangle, inner sep=5pt, right=of letter3] {k};

  \begin{pgfonlayer}{background}
    \node[draw, rounded corners, fill=blue!10, inner sep=10pt,
          fit=(letter1) (letter2) (letter3) (letter4),
          label=above:{\textbf{Input Text}}] (inputbox) {};
  \end{pgfonlayer}
  
  \node (vector1) [draw, rectangle, inner sep=5pt, below=1.5cm of letter1] 
    {$\displaystyle \begin{pmatrix}0.2\\ \vdots\\0.8\end{pmatrix}$};
  \node (vector2) [draw, rectangle, inner sep=5pt, below=1.5cm of letter2] 
    {$\displaystyle \begin{pmatrix}0.3\\ \vdots\\0.9\end{pmatrix}$};
  \node (vector3) [draw, rectangle, inner sep=5pt, below=1.5cm of letter3] 
    {$\displaystyle \begin{pmatrix}0.5\\ \vdots\\0.6\end{pmatrix}$};
  \node (vector4) [draw, rectangle, inner sep=5pt, below=1.5cm of letter4] 
    {$\displaystyle \begin{pmatrix}0.7\\ \vdots\\0.1\end{pmatrix}$};
  
  \begin{pgfonlayer}{background}
    \node[draw, rounded corners, fill=yellow!10, inner sep=10pt,
          fit=(vector1) (vector2) (vector3) (vector4)] (embedbox) {};
  \end{pgfonlayer}
  
  \begin{pgfonlayer}{foreground}
    \node[draw, fill=white, inner sep=1pt] at ([yshift=2mm]embedbox.north) {\textbf{Vector Embedding}};
  \end{pgfonlayer}
  
  \draw[->] (letter1) -- (vector1);
  \draw[->] (letter2) -- (vector2);
  \draw[->] (letter3) -- (vector3);
  \draw[->] (letter4) -- (vector4);
  
  \draw[->, thick]
    ($(vector1.south west)+(0,-0.5)$) -- node[below, font=\footnotesize]{Time}
    ($(vector4.south east)+(0,-0.5)$);
  
  \node (reservoir) [draw, circle, fill=orange!40, minimum size=3cm, right=0.5cm of embedbox, align=center] 
    {\textbf{Reservoir} \\ \textbf{Att-Enhanced Reservoir} \\ \textbf{Transformer}};
  
  \draw[->, thick] (embedbox.east) .. controls +(1,0) and +(-1,0) .. (reservoir.west);
  
  \node (outweights) [draw, rectangle, rounded corners, fill=purple!30, inner sep=10pt,
                      minimum width=2.5cm, minimum height=3cm, right=0.5cm of reservoir,
                      label=above:{\textbf{Output Layer}}] {
    $\displaystyle \begin{pmatrix}w_1\\w_2\\\vdots\\w_m\end{pmatrix}$
  };
  
  \draw[->, thick] (reservoir) -- (outweights);
  
  \node (pred) [draw, rectangle, rounded corners, fill=green!30, inner sep=5pt,
                minimum width=2.5cm, minimum height=3cm, right=0.5cm of outweights] {
    $\displaystyle \begin{pmatrix}p_a=0.1\\ p_b=0.01 \\ p_c=0.12\\ \vdots\end{pmatrix}$
  };
  
  \draw[->, thick] (outweights) -- (pred);
  
  \node (loss) [draw, rectangle, rounded corners, fill=red!10, inner sep=5pt, above=0.5cm of pred,
               label=above:{\textbf{Cross Entropy – Prediction}}] {
    $\displaystyle H(y,\hat{y}) = -\sum_{i} y_i \log \hat{y}_i$
  };
  
  \draw[->, thick, dashed] (loss) -- (pred);
  
\end{tikzpicture}
\caption{Diagram illustrating all three ML agent models with vector embedding, processing (Reservoir, AERC, Transformer), output layer, and predicted probability vector for the next letter. The vector embedding is trained for the transformer case, while being randomly initialized for the reservoirs. The predicted probabilities (e.g., $p_A$ for letter a) are used to compute the cross entropy loss $H(y,\hat{y})=-\sum_i y_i\log\hat{y}_i$, which drives the error feedback.}
\label{fig:reservoir}
\end{figure*}

\emph{Reservoir computing} (RC) offers an alternative computational paradigm. In RC, a large, fixed recurrent ``reservoir'' projects inputs into a high-dimensional state space, and only a lightweight readout layer is trained. This design dramatically reduces training time and energy consumption and can be implemented efficiently on analog hardware substrates. It has also been shown to excel in time-series prediction \cite{jaeger2004harnessing, pathak2018model, krishnagopal2020separation}, capitalizing on its large, fixed random recurrent structures for capturing nonlinear dynamics with minimal training overhead \cite{jaeger2001echo, lukosevicius2009reservoir, maass2002real, schrauwen2007overview}. A key advantage of reservoir computing lies in its versatile physical realizations, spanning electronic, optical, and even quantum hardware \cite{fernando2003bucket, ghosh2019quantum, negoro2018machine, chen2020temporal}.

Recently, more expressive control over reservoir behavior has been introduced via neural programming paradigms, further expanding their utility in symbolic computation and structured sequence modeling \cite{kim2022neural}, while other approaches introduced attention mechanisms at the readout stage to adaptively weight reservoir states \cite{koester2024attention}. While prior work has explored the use of reservoirs for text classification tasks \cite{schaetti2019behaviors} and even demonstrated hardware-oriented implementations for language learning \cite{sun2021insensor}, the use of reservoir computing explicitly as a \emph{language model}—that is, for generative next-token prediction—remains largely unexplored.

Although transformers remain the dominant and most effective architecture for large-scale Natural Language Processing (NLP) \cite{brown2020language}, our goal is not to challenge this paradigm but to explore a complementary direction. Reservoir computing offers fundamentally different computational characteristics: fixed high-dimensional recurrent dynamics, strong energy efficiency, and inherent suitability for analog, neuromorphic, or photonic substrates. In settings where computational or energy budgets are constrained or full training of the underlying hardware is not feasible, such as embedded or edge devices, these properties make RC an appealing alternative. To reflect this motivation more clearly, the present work is framed as a proof-of-concept study investigating whether reservoir-based architectures can support next-token language modeling.

In this study, we present a unified framework, applying three distinct approaches — namely the classic RC, the attention-enhanced RC, and the transformer approach — to \emph{character-level} sequence prediction on a small corpus of Shakespeare text inspired by recent minimalist transformer implementations such as Karpathy's NanoGPT \cite{karpathy2022nanogpt}. We vary reservoir size in the classic RC and RC size and attention layer size for an attention-enhanced readout \cite{koester2024attention}, while varying hidden dimension and number of layers in transformers, obtaining multiple configurations spanning from thousands up to hundred thousand of trainable parameters. By measuring cross-entropy loss across these diverse setups, we expose fundamental trade-offs in resource usage and generalization performance.

In this work we additionally introduce a layered reservoir language model that adapts reservoir computing to modern token-level causal language modeling. By stacking multiple fixed recurrent reservoirs together with lightweight trainable mixing and feed-forward components, the model retains the depth and normalization structure characteristic of transformer architectures while replacing self-attention layers with computationally efficient recurrent dynamics. We evaluate this architecture at various model scales, compare it against transformer baselines under matched parameter budgets, and analyze both optimization behavior and scaling characteristics.

\section{Methods and Model Architectures}

\subsection{Traditional Reservoir Computing}

The echo state implementation of a classic reservoir computer consists of three main components: an input-to-reservoir weight matrix $\mathbf{W}_{\mathrm{in}} \in \mathbb{R}^{N \times d}$, a fixed recurrent reservoir matrix $\mathbf{W}_{\mathrm{res}} \in \mathbb{R}^{N \times N}$, and a readout layer $\mathbf{W}_{\mathrm{out}} \in \mathbb{R}^{\mathcal{V} \times N}$ \cite{jaeger2001echo}. Here $d$ is the dimension of the chosen input-embedding vector for the input letter, $N$ the number of nodes in the reservoir, and $\mathcal{V}$ the size of the vocabulary. We treat all state and input vectors as column vectors. The reservoir evolves its hidden states $\mathbf{r}_t \in \mathbb{R}^{N}$ according to
\begin{align}
    \mathbf{r}_t = \tanh\!\big(\mathbf{W}_{\mathrm{res}}\,\mathbf{r}_{t-1} + \mathbf{W}_{\mathrm{in}}\,\mathbf{x}_t\big),
\end{align}
where $\mathbf{x}_t \in \mathbb{R}^{d}$ is the embedding vector of the current input character. In the simplest (\emph{traditional}) reservoir setting, only the readout layer $\mathbf{W}_{\mathrm{out}}$ is trained:
\begin{align}
    \bar{\mathbf{y}}_t = \mathbf{W}_{\mathrm{out}}\,\mathbf{r}_t,
\end{align}
where \( \bar{\mathbf{y}}_t \in \mathbb{R}^{\mathcal{V}} \) denotes the unnormalized logits at time \( t \). Typically, \( \mathbf{W}_{\mathrm{out}} \) is learned using ridge regression, which minimizes the mean squared error with an added \( \ell_2 \) regularization term to prevent overfitting and improve generalization. However, in our simulations, we employ a cross-entropy loss function, necessitating gradient-based optimization methods for training \( \mathbf{W}_{\mathrm{out}} \). Moreover, ridge regression can become computationally intensive as the number of training samples and reservoir size increase, due to the inversion of large matrices. By using gradient-based methods, we circumvent these scalability issues.

\begin{table*}[ht]
    \centering
    \caption{Trainable Parameter Counts for the classic reservoir, AERC, and Transformer Architectures. All systems used an embedding dimension of 16 for the 59 Vocabularies.}
    \label{tab:parameter_counts}
    \begin{tabular}{|l|l|c|}
        \hline
        \textbf{Approach} & \textbf{Configuration} & \textbf{Parameters} \\ \hline
        \multirow{5}{*}{Transformer} 
            & Hidden Size $d_h=64$, Heads $h=4$, Layers $L=4$ 
            & $15067$ \\ \cline{2-3}
            & Hidden Size $d_h=72$, Heads $h=8$, Layers $L=8$ 
            & $30299$ \\ \cline{2-3}
            & Hidden Size $d_h=128$, Heads $h=8$, Layers $L=8$ 
            & $45083$ \\ \cline{2-3}
            & Hidden Size $d_h=356$, Heads $h=8$, Layers $L=8$ 
            & $105275$ \\ \cline{2-3}
            & Hidden Size $d_h=256$, Heads $h=16$, Layers $L=16$ 
            & $155803$ \\ \hline
        \multirow{5}{*}{Reservoir}
            & Reservoir Size $N=250$ 
            & $14809$ \\ \cline{2-3}
            & Reservoir Size $N=500$ 
            & $29559$ \\ \cline{2-3}
            & Reservoir Size $N=750$ 
            & $44309$ \\ \cline{2-3}
            & Reservoir Size $N=1750$ 
            & $103309$ \\ \cline{2-3}
            & Reservoir Size $N=2600$ 
            & $153459$ \\ \hline
        \multirow{5}{*}{Att Reservoir}
            & Reservoir Size $N=75$, Hidden Size $H=13$ 
            & $15464$ \\ \cline{2-3}
            & Reservoir Size $N=75$, Hidden Size $H=19$ 
            & $31124$ \\ \cline{2-3}
            & Reservoir Size $N=100$, Hidden Size $H=20$ 
            & $45259$ \\ \cline{2-3}
            & Reservoir Size $N=150$, Hidden Size $H=25$ 
            & $102809$ \\ \cline{2-3}
            & Reservoir Size $N=160$, Hidden Size $H=30$ 
            & $155459$ \\ \hline
    \end{tabular}
\end{table*}

\subsection{Attention-Enhanced Reservoir Computing}
Incorporating an attention mechanism into a RC can significantly improve its performance by making the output weights adaptable to the reservoir states, although increasing the number of trainable weights relative to the reservoir size \cite{koester2024attention}. Instead of using fixed output weights as in traditional RC, the attention-enhanced reservoir computing (AERC) uses dynamically computed attention weights that vary with inputs.

For each data point \( l \) fed into the reservoir, the corresponding attention weights \( \mathbf{W}_{\text{att}, l} \in \mathbb{R}^{H_o \times N} \) are calculated. These attention weights are derived using a small neural network \( F \) with trainable parameters \( \mathbf{W}_{\text{net}} \), based on the reservoir states \( \mathbf{r}_l \):
\begin{align}
\mathbf{W}_{\text{att}, l} = F(\mathbf{W}_{\text{net}}, \mathbf{r}_l).
\end{align}
The weights are used to project the reservoir state onto a small hidden state $\mathbf{r_o}_l \in \mathbb{R}^{H_o}$:
\begin{align}
\mathbf{r_o}_l = \mathbf{W}_{\text{att},l} \mathbf{r}_l.
\end{align}
This intermediate output vector is then mapped onto the final logits of dimensions $\mathcal{V}$ via a final linear matrix $\mathbf{W}_{out} \in \mathbb{R}^{\mathcal{V} \times H_o}$:
\begin{align}
\bar{\mathbf{y}}_l = \mathbf{W}_{out} \mathbf{r_o}_l.
\end{align}
This choice of architecture with an intermediate layer is done to reduce the number of parameters.

Our goal is to refine \(\mathbf{W}_{\mathrm{net}}\) so that the AERC can effectively model complex temporal dependencies in the input data. In this work, the neural network component consists of a single hidden layer with a ReLU activation function.
After the training phase, the attention mechanism computes new attention weights at each time step. These dynamically updated attention weights allow the system to adapt to the evolving input, improving performance in complex dynamical systems.

\subsection{Transformer Architectures}
Transformers \cite{vaswani2017attention} are fully trainable neural networks that have achieve state-of-the-art performance on many sequence modeling tasks. The input sequence of characters is mapped to embeddings of dimension $d$. A stack of $L$ layers follows, each containing a multi-head self-attention block and a feed-forward sub-block. Attention heads learn to focus on relevant parts of the input sequence at different positions. The feed-forward layers, typically parameterized by matrices of dimension on the order of $d_h = d \times \alpha d$ (where $\alpha$ is a constant expansion factor), refine the representation. The complexity of the transformer grows with $L$ (the number of layers) and $d$ (the embedding dimension). The output is passed through a final projection to logits of size $\mathcal{V}$, the vocabulary dimension. All parameters (including embedding, attention, feed-forward, and output projection) are learned end-to-end, giving transformers a large capacity to capture long-range dependencies. However, this capacity often comes at a significant computational and memory cost, particularly for large $L$, $d$ and long sequences.

In standard transformer architectures, the expansion factor $\alpha$ is typically chosen between $2$ and $4$, making the intermediate feed-forward dimension $d_{h} = \alpha d$ larger than the embedding dimension. In our simulations, however, we keep the embedding dimension fixed at $d=16$ and vary only the internal feed-forward dimensions to ensure that all architectures—Transformer, classic reservoir, and AERC—operate under comparable embedding dimensions. This design choice enables a fair comparison of representational efficiency across models without conflating performance differences with changes in embedding dimensionality.

\begin{figure*}[t]
    \centering
    \includegraphics[width=\textwidth]{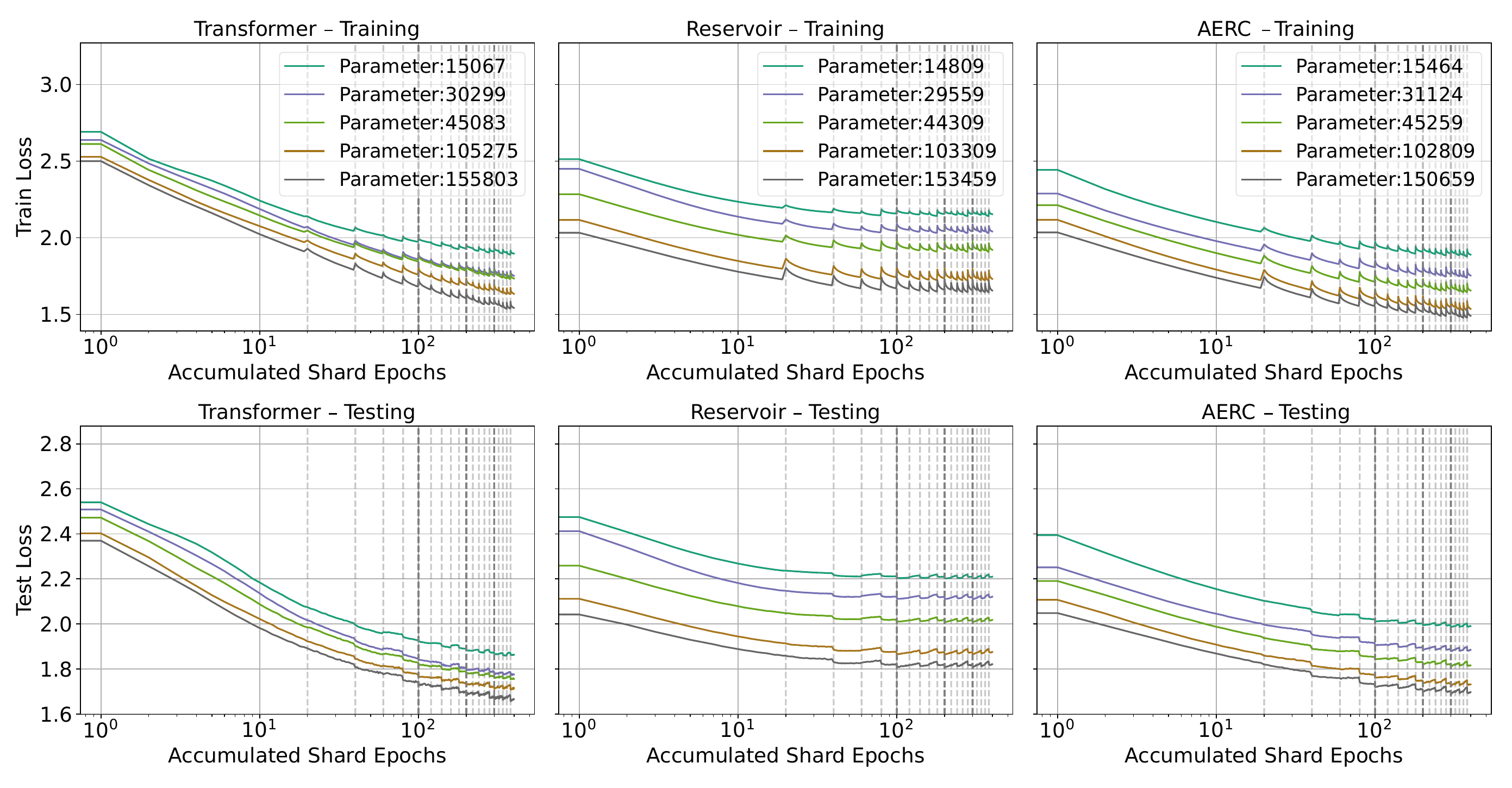}  
    \caption{Training and testing loss over accumulated number of shared epochs for the Transformer, Reservoir, and Attention-Enhanced Reservoir Computer for the 5 different complexity models of Table \ref{tab:parameter_counts}. The thin light-grey dashed vertical lines indicate a shard change, while the black dashed lines show one complete epoch of all shards. Every shard was run for 5 shard epochs until the next one was processed.}
    \label{fig:training_testing_loss}
\end{figure*}

\subsection{Architectures}

The number of trainable parameters depends strongly on the chosen architecture and hyperparameters. We focused on choosing hyperparameters, such that the number of trainable parameters is comparable between all 3 different models.
Table~\ref{tab:parameter_counts} summarizes the number of trainable parameters for each architecture across different configurations. With a vocabulary size of \(V = 59\) characters due to the Shakespeare dataset used in our simulations we arrive at architectures that range from around 15000 trainable parameters to around 150000. \\

For all three models, we take an embedding dimension of $d = 16$.
For the Transformer model we vary the hidden size $d_h$, the number of heads $h$, and the number of layers $L$.
For the classic reservoir and AERC configurations, we vary the reservoir size $N$, the hidden dimensions of the attention layer $H$, and set the final layer $H_o=H$ to be equal to the attention layers hidden dimension.
In the reservoir-based architectures (classic reservoir and AERC), only the readout and gating networks are trained using gradient descent. The internal reservoir matrices $W_{\mathrm{in}}$ and $W_{\mathrm{res}}$ are initialized once and remain fixed throughout training; they are therefore not counted as trainable parameters in Table~\ref{tab:parameter_counts}. This definition is consistent with the long-term perspective of reservoir computing, where the reservoir dynamics are viewed as a fixed free computation implemented directly in analog or photonic hardware—while only digital components are optimized.

\subsection{Training}

We train the parameters of all three architectures to predict the next character in a sequence by minimizing the cross-entropy loss:
\begin{align}
    \mathbf{W}_{\mathrm{net}}^{(s+1)} = \mathbf{W}_{\mathrm{net}}^{(s)} - \gamma \nabla F\big(\mathbf{W}_{\mathrm{net}}^{(s)}, \mathbf{R}\text{/}\mathbf{X}\big),
\end{align}
where \(s\) indexes the training batches, \(\gamma\) is the learning rate, and in the two reservoir computer cases \(\mathbf{R}\) denotes the reservoir states and in the Transformer case \(\mathbf{X}\) the input sentence embedding. One batch corresponds to a subset of the training data used in each update step. The cross-entropy loss is defined as
\begin{align}
    H(y,\hat{y}) = -\sum_{i}^\mathcal{V} y_i \log \hat{y}_i    
    \label{eq:CE}
\end{align}
where \(\mathcal{V}\) is the number of output classes, \(\hat{y}_i\) are the output probabilities, and \(y_{i}\) is the true one-hot encoded label.
We apply the softmax function to obtain output probabilities and compute the cross-entropy loss accordingly, a standard formulation in classification tasks \cite{bishop2006pattern}.
All trainable weights are updated using backpropagation \cite{rumelhart1986learning} and optimized using the Adam optimizer, a widely adopted variant of stochastic gradient descent \cite{kingma2015adam}.

\subsection{Dataset and Methodology}

We study a character-level next-character prediction task. Let
$\mathbf{X}=\{x_1,\dots,x_T\}$ denote a fixed-length input sequence of characters, and let
$y_{T+1}$ be the next character to predict. The task is formulated as a multi-class
classification problem over a vocabulary of size $\mathcal{V}=59$.

We use a 9-million-character corpus of Shakespeare text, inspired by minimalist
transformer implementations such as Karpathy's NanoGPT~\cite{karpathy2022nanogpt}.
All text is lowercased during preprocessing to reduce the vocabulary size.
The corpus is split into six contiguous shards: five shards for training and one shard
for testing. Each shard is processed at the character level, yielding sequences of
length $32$ as input, with the subsequent character serving as the prediction target.

Training minimizes the standard cross-entropy loss between predicted logits and the
ground-truth next character. Models are trained using Adam with a learning rate of
$1\times10^{-4}$ and a batch size of $1024$. We track both training and test losses per
shard and per epoch, reporting cross-entropy on the held-out test shard as the primary
performance metric.

All models—the traditional reservoir, the attention-enhanced reservoir, and the
transformer baselines—are trained end-to-end via backpropagation within a unified
training framework. In addition to next-character prediction accuracy, we evaluate
closed-loop text generation by feeding predicted characters back as input and analyzing
distributional statistics and repetitive patterns. Our parameter sweep varies the
reservoir size or neural subnetwork size in the attention-based reservoir models, as
well as the hidden dimension $d$ and number of layers $L$ (and heads) in the transformer
models.

To reduce computational cost, reservoir responses are precomputed shard-wise and kept
in memory, allowing several training epochs to be run on each shard before advancing to
the next. This process is repeated cyclically across all shards.

All simulations were conducted on a single NVIDIA RTX~4090 GPU running Ubuntu~22.04,
using PyTorch~2.4.1 with CUDA~12.1. Mixed-precision training (bf16/fp16 with loss
scaling) was used throughout. The full codebase, configuration files, and simulation
scripts are publicly available at (See Data Availability).

\section{Results}

\subsection{Comparison of Training and Testing Loss}

We now present the performance of the trained models and analyze the training process.  

To quantitatively assess model performance, we report training and testing losses across epochs and data shards for all five model sizes described in Table~\ref{tab:parameter_counts}.  
The results are visualized in Figure~\ref{fig:training_testing_loss}.
Figure~\ref{fig:training_testing_loss} top row presents training  and bottom row testing losses for the five model sizes listed in Table~\ref{tab:parameter_counts}, ordered from smallest (turquoise) to largest (grey), where the first column shows the transformer, the second column the reservoir, and the third column the AERC.
Model sizes range from approximately 15{,}000 to 155{,}000 trainable parameters.

Training is performed in \emph{shards}: for each shard, reservoir responses are precomputed and stored in memory. Multiple training epochs are then performed on each shard before proceeding to the next.  
This approach mimics the inherent speed of reservoir computing, which theoretically operates with near-instantaneous dynamics.  
Vertical light-gray dashed lines mark shard transitions, while thicker dark-gray dashed lines denote full passes over the dataset (epochs).  
Due to this sharding structure, we observe periodic spikes in both training and testing losses—particularly in training—caused by model fine-tuning on individual shards.  
While increased stochasticity might reduce overfitting and improve convergence by avoiding poor local minima, our configuration of six shards and five epochs per shard balances computational efficiency and generalization.  
This setup tries to maximize the benefits of precomputing reservoir states while mitigating overfitting through moderate stochasticity.

\begin{figure}[t]
    \centering
    \includegraphics[width=0.5\textwidth]{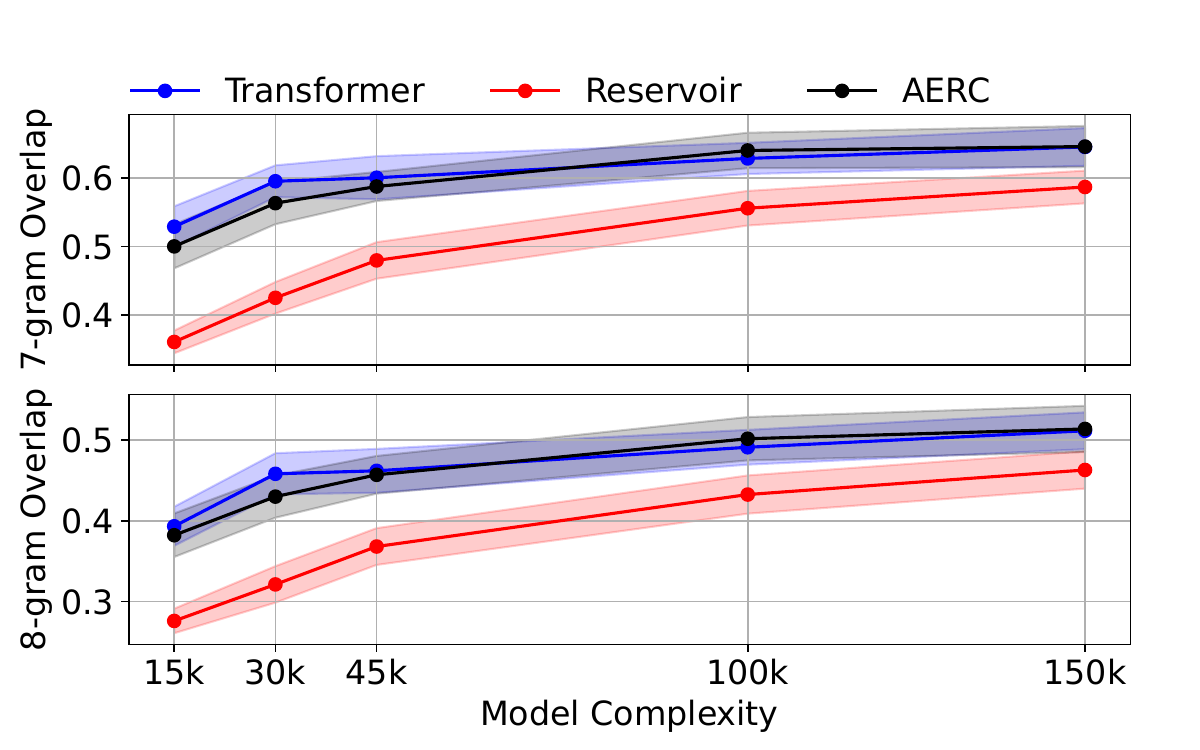}  
    \caption{7-gram and 8-gram overlap for the Transformer, Reservoir, and Attention-Enhanced Reservoir Computer over the number of trainable parameters.}
    \label{fig:ngram_overlap}
\end{figure}

Analyzing the loss curves for the Transformer, the classical Reservoir Computer (RC), and the Attention-Enhanced Reservoir Computer (AERC), we observe that loss variance across different model sizes is more pronounced in the RC and AERC than in the Transformer.  
As expected, the Transformer achieves the best overall performance, with a minimum test loss of 1.67 for the largest configuration.  
However, we note that the classical RC achieves a test loss as low as 1.81, and the AERC reaches an intermediate value of 1.73—indicating that both RC-based models are competitive despite their relative simplicity.

Finally, the Transformer exhibits the smallest training–testing loss gap, reflecting better generalization and reduced overfitting.  
In contrast, both reservoir-based models show a greater tendency to overfit, likely due to the fixed nature of their reservoirs and the limited capacity of their readout layers.

\begin{figure*}[t]
    \centering
    \includegraphics[width=\textwidth]{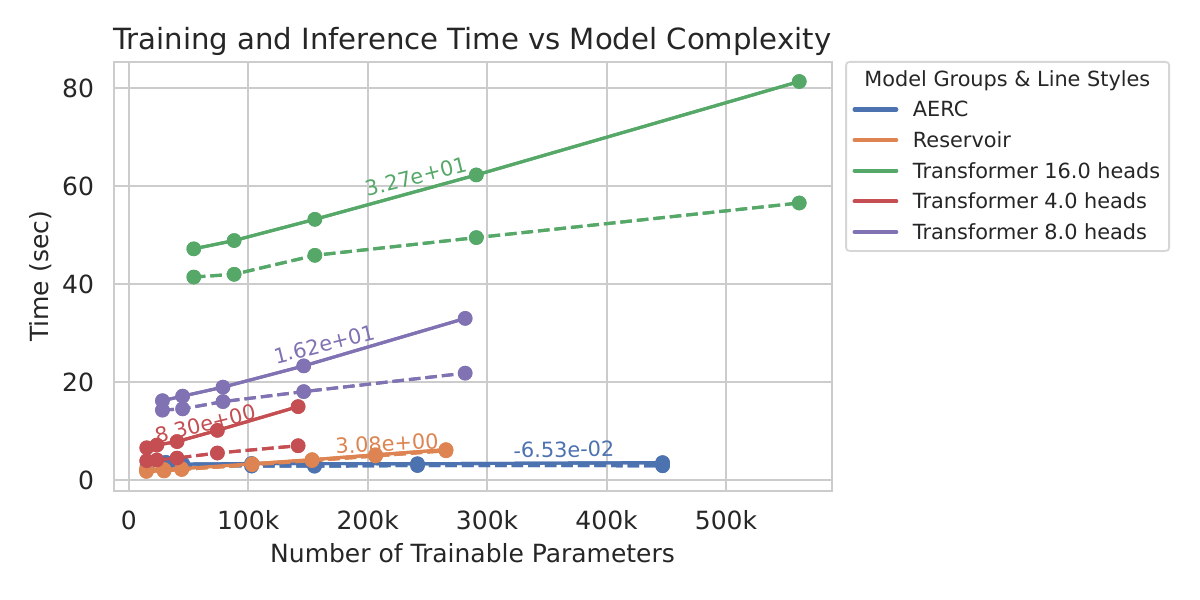}  
    \caption{Training (solid) and inference (dashed) time for a longer text generation over the number of trainable parameters for the Transformer, Reservoir, and Attention-Enhanced Reservoir Computer. The numbers above the line are from a fit applied via $y=\alpha\log_{10}(x)$, where $x$ is the number of trainable parameters, $y$ is the training time and $\alpha$ the slope.}
    \label{fig:time}
\end{figure*}

\subsection{Evaluation Metrics and Analysis}
\label{sec:evaluation_metrics}
Next we investigate the long-term text similarity performance of the three architectures for auto-regressive generation.
To assess the performance of the different model architectures in generating longer text, we employ the N-gram overlap to quantify performance beyond the loss of cross entropy. This metric quantifies the aspect of similarity to the reference data and distributional alignment and is inspired by classical N-gram-based evaluation metrics such as BLEU \cite{papineni2002bleu}, which assess the proportion of overlapping sequences between generated and reference text. The following subsection details the metric.

\textit{N}-gram Overlap measures the similarity between the generated text and a reference text by computing the fraction of unique \textit{n}-grams in the generated text that also appear in the reference text.

\begin{figure*}[t]
    \centering

    \tikzset{
        laercblock/.style={draw, rounded corners, thick, align=center,
                           minimum height=1.4em, minimum width=7.0em, inner sep=4pt},
        smallblock/.style={draw, rounded corners, thick, align=center,
                           minimum height=1.1em, minimum width=7.0em, inner sep=3pt},
        line/.style={-Stealth, thick},
        brace/.style={decorate, decoration={brace, amplitude=5pt}, thick},
        resid/.style={-Stealth, thick, dashed},
    }

    \begin{subfigure}{0.48\linewidth}
    \centering
    \begin{tikzpicture}[node distance=0.9cm]
        \node[laercblock] (tokens) {Input\\tokens};
        \node[laercblock, below=0.45cm of tokens] (embed) {Token\\embedding};
        \node[laercblock, below=0.45cm of embed] (posenc) {Positional\\encoding};
        \node[laercblock, below=0.45cm of posenc] (sum) {Embedding $+$\\position};

        \node[smallblock, below=0.55cm of sum] (rb1) {LAERC block 1};
        \node[smallblock, below=0.40cm of rb1] (rb2) {LAERC block 2};
        \node[smallblock, below=0.40cm of rb2] (rbL) {LAERC block $L$};
    
        \node[draw, rounded corners, thick, inner sep=4pt,
              fit=(rb1) (rbL)] (stackbox) {};
    
        \draw[brace] ([xshift=0.35cm]rb1.north east) --
                     ([xshift=0.35cm]rbL.south east)
                     node[midway,xshift=0.45cm,rotate=270,anchor=center]
                     {\scriptsize $L$ stacked blocks};
    
        \node[laercblock, below=0.55cm of rbL] (lnf) {Final\\LayerNorm};
        \node[laercblock, below=0.55cm of lnf] (logits) {Linear\\projection};
        \node[laercblock, below=0.55cm of logits] (softmax) {Softmax\\(next token)};
    
        \draw[line] (tokens) -- (embed);
        \draw[line] (embed) -- (posenc);
        \draw[line] (posenc) -- (sum);
        \draw[line] (sum) -- (rb1);
        \draw[line] (rb1) -- (rb2);
        \draw[line, dashed] (rb2) -- (rbL); 
        \draw[line] (rbL) -- (lnf);
        \draw[line] (lnf) -- (logits);
        \draw[line] (logits) -- (softmax);
    \end{tikzpicture}
    \caption{LAERC language model.}
    \end{subfigure}
    \hfill
    \begin{subfigure}{0.48\linewidth}
    \centering
    \begin{tikzpicture}[node distance=0.7cm]
    \node[smallblock] (xin) {$\mathbf{x}_t^{(\ell)}$};
    \node[smallblock, below=0.55cm of xin] (lnres) {LayerNorm$_{\text{res}}$};
    \node[smallblock, below=0.55cm of lnres] (res) {Fixed reservoir RNN};
    \node[smallblock, below=0.55cm of res] (lns) {LayerNorm$_{\text{res,out}}$};
    \node[smallblock, below=0.55cm of lns] (resmlp) {Res-MLP $\rightarrow \mathbf{r}_t^{(\ell)}$};
    \node[smallblock, below=0.55cm of resmlp] (mixln) {Concat\ \&\ LayerNorm$_{\text{mix}}$};
    \node[smallblock, below=0.55cm of mixln] (gate) {Gate $\mathbf{g}_t^{(\ell)}$};
    \node[smallblock, below=0.55cm of gate] (mix) {Mix $\mathbf{m}_t^{(\ell)}$};
    \node[smallblock, below=0.55cm of mix] (lnffn) {LayerNorm$_{\text{ffn}}$};
    \node[smallblock, below=0.55cm of lnffn] (ffn) {FFN};
    \node[smallblock, below=0.55cm of ffn, text width=7.6em]
          (rezero) {ReZero};

    \draw[line] (xin) -- (lnres);
    \draw[line] (lnres) -- (res);
    \draw[line] (res) -- (lns);
    \draw[line] (lns) -- (resmlp);
    \draw[line] (resmlp) -- (mixln);
    \draw[line] (mixln) -- (gate);
    \draw[line] (gate) -- (mix);
    \draw[line] (mix) -- (lnffn);
    \draw[line] (lnffn) -- (ffn);
    \draw[line] (ffn) -- (rezero);

    \draw[resid] (xin.west) to[out=180,in=180] (mixln.west);
    \draw[resid] (mix.west) to[out=180,in=180] (rezero.west);

    \node[draw, rounded corners, thick, inner sep=6pt,
          fit=(lnres) (res) (lns) (resmlp)] (respathbox) {};

    \draw[brace] ([xshift=0.45cm,yshift=0.1cm]respathbox.north east) --
                 ([xshift=0.45cm,yshift=-0.1cm]respathbox.south east)
                 node[midway,xshift=0.55cm,rotate=270,anchor=center]
                 {\scriptsize LAERC path};

    \node[draw, rounded corners, thick, inner sep=6pt,
          fit=(lnffn) (ffn)] (ffnbox) {};

    \draw[brace] ([xshift=0.45cm,yshift=0.1cm]ffnbox.north east) --
                 ([xshift=0.45cm,yshift=-0.1cm]ffnbox.south east)
                 node[midway,xshift=0.55cm,rotate=270,anchor=center]
                 {\scriptsize FFN};
\end{tikzpicture}
\caption{Single LAERC block.}
    \end{subfigure}

    \caption{Overview of the layered attention-enhanced reservoir computing (LAERC)
    architecture: (a) the full language model with $L$ stacked reservoir blocks and
    (b) the internal structure of a single block combining a fixed reservoir,
    gating, and ReZero feed-forward refinement. The dashed lines indicate residual connections, enabling the LAERC to backpropagate past the fixed unknown reservoir.}
    \label{fig:laerc_arch}
\end{figure*}

\begin{align}
    \text{Overlap-n} = \frac{\left| \left\{G_n \cap R_n \right\} \right|}{\left| G_n \right|},
\end{align}

\noindent
where $G_n$ is the set of unique \textit{n}-grams in the generated text, and $R_n$ is the set of unique \textit{n}-grams in the reference (test) text.

This metric assesses how well the model captures the \textit{n}-gram patterns present in the reference data. A higher overlap indicates that the generated text shares more common sequences with the reference.

\subsection{\textit{N}-Gram Similarity}
The results of the 7-gram and 8-gram performance are shown in Fig. \ref{fig:ngram_overlap}.
The evaluation metrics reveal that the classic reservoir approach tends to have the lowest overlap and thus the lowest generalization. 
Transformers and Attention-Enhanced Reservoirs demonstrate on par diversity and distributional alignment. Their ability to maintain higher performance underscores their superior performance, due to the inclusion of neural networks.

The traditional reservoir, though the simplest, can struggle with capturing complex dependencies and maintaining distributional alignment, as evidenced by its lower distinct-\textit{n} scores. However, its simplicity remains valuable in scenarios where an easy hardware implementation is prioritized.

The attention-enhanced reservoir improves upon the traditional model by incorporating dynamic readout weights, thereby enhancing its ability to capture more complex patterns.

\begin{figure*}[t]
    \centering
    \includegraphics[width=0.85\linewidth]{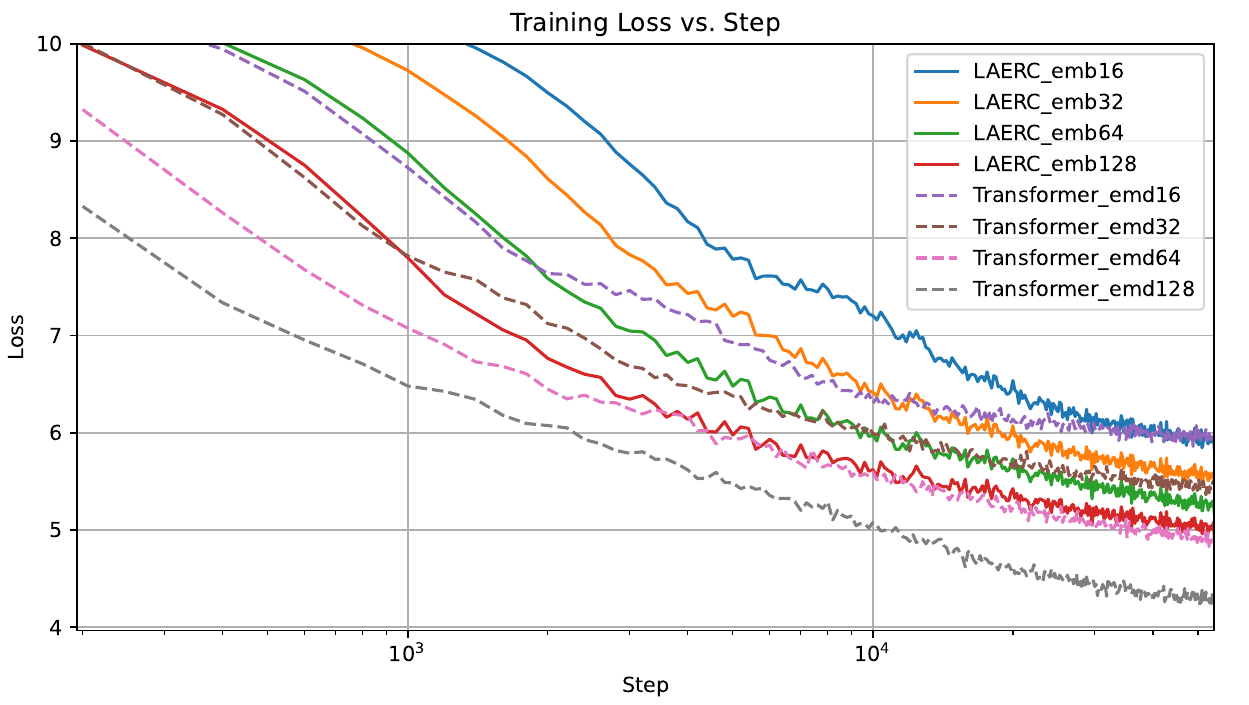}
    \caption{Smoothed training loss as a function of the global optimization step for transformer baselines and layered reservoir (ReservoirFFN) language models trained on OpenWebText. Each curve corresponds to a different embedding dimension and thus a different total parameter count.}
    \label{fig:layered_loss}
\end{figure*}

\subsection{Inference Time Comparison}

As a final performance metric, we evaluate the training and inference time costs for all three model architectures.  
Figure~\ref{fig:time} displays the time (in seconds) required for both training and inference, plotted against the number of trainable parameters in each model.  
Inference time refers to generating a sequence over a fixed data chunk or producing longer text samples.

Notably, we exclude the computational cost of the reservoir dynamics themselves. 
We assume that the reservoir computation can be offloaded to physical substrates such as photonic circuits, which have been experimentally shown to operate at extremely high speeds \cite{vandoorne2014experimental}.
Such hardware would operate at time scales several orders of magnitude faster than the subsequent postprocessing by the trained layers, justifying its omission in this comparison.

Furthermore, although transformer architectures benefit from high degrees of parallelization in digital hardware, the sequential nature of reservoir dynamics does not impose a practical bottleneck when implemented on physical substrates. Experimental demonstrations using photonic and analog reservoirs show operational speeds over GHz the internal state evolution occurs effectively instantaneously compared to the digitally executed postprocessing stages \cite{brunner2013parallel, vandoorne2014experimental}.
Therefore, physical reservoirs can be regarded as \emph{practically parallel} from the perspective of overall system throughput. In contrast, transformer attention layers remain constrained by digital memory bandwidth and quadratic scaling with respect to sequence length.

In Figure~\ref{fig:time}, solid lines indicate training time, while dashed lines represent inference time. 
The color scheme distinguishes between the three model types: classic reservoir (orange), attention-enhanced reservoir (blue), and transformer (green, red, and purple).   
It is evident that reservoir-based models exhibit substantially lower computational times, and their scalability with respect to parameter count is more favorable than that of transformers.
The numbers above the line are from a fit applied via $y=\alpha\log_{10}(x)$, where $x$ is the number of trainable parameters, $y$ is the training time and $\alpha$ the slope.
From the graphs itself and the slope $\alpha$ we can see that reservoir based training and inference time is by magnitudes of order faster.
This performance advantage arises because the reservoir handles the computationally expensive sequence processing, whereas in transformers, attention mechanisms scale quadratically with sequence length.  
These results suggest that using a reservoir for data preprocessing can significantly accelerate language model performance without incurring the high time cost associated with full transformer architectures.

\section{Layered Reservoir Language Model}
\label{sec:layered_reservoir_lm}

While the previous sections focused on single-reservoir and attention-enhanced reservoir architectures for character-level language modeling with smaller models for edge-computing scenarios, we now extend this framework to a \emph{layered} reservoir language model designed for large-scale token-level training. The central idea is to replace the quadratic-complexity self-attention layers of transformers with fixed recurrent reservoirs, while retaining lightweight trainable components for representation refinement, gating, and output projection.

We refer to this architecture as \emph{Layered Attention-Enhanced Reservoir Computing (LAERC)}. Each layer contains a fixed, untrained recurrent reservoir that performs temporal mixing, combined with trainable feed-forward and gating components that adaptively integrate reservoir-induced features into the model’s representation. Residual connections allow gradients to bypass the fixed reservoirs, enabling end-to-end training using standard backpropagation despite the reservoirs themselves being untrainable.

\subsection{Architecture}

A token sequence is first mapped to a sequence of continuous embeddings in the same manner as standard transformer language models. These embeddings are then processed by a stack of $L$ identical reservoir blocks, each operating sequentially. A self-contained technical definition can be found in Sec. \ref{sec:app_1}.

Each LAERC block consists of four conceptual components: a fixed recurrent reservoir, a small trainable MLP that projects reservoir states back into the model dimension, an input-dependent mixing gate, and a feed-forward refinement stage. The overall structure is illustrated in Fig.~\ref{fig:laerc_arch}.

\begin{figure*}[t]
    \centering
    \includegraphics[width=0.85\linewidth]{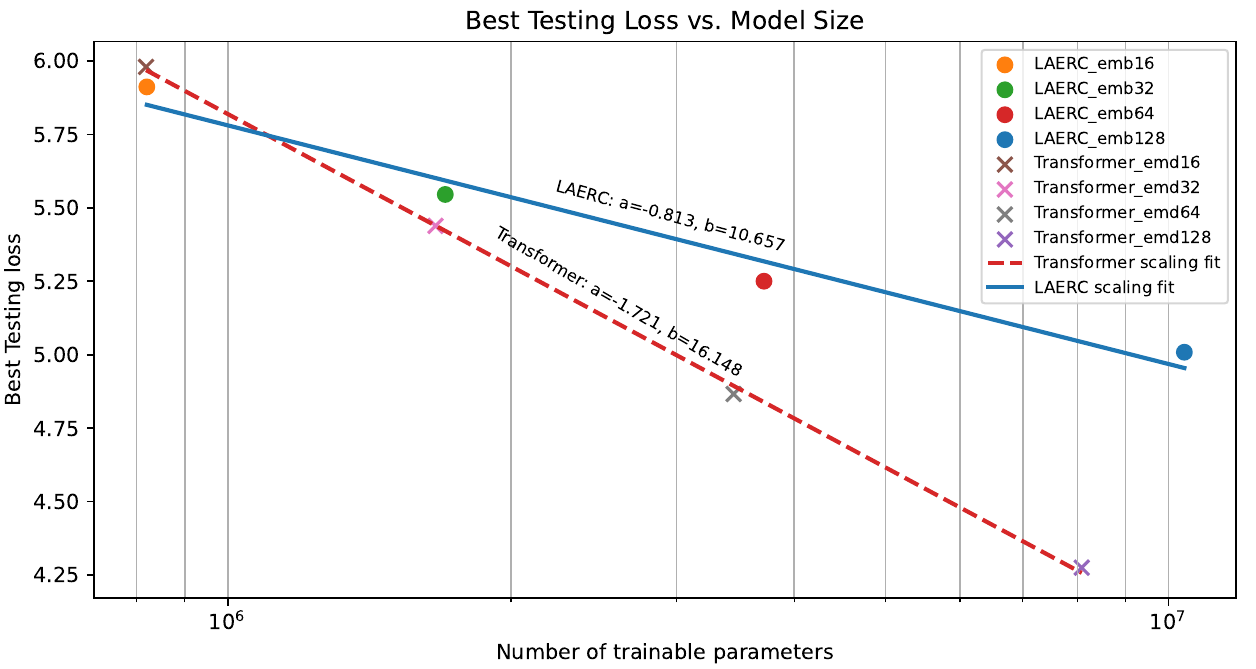}
    \caption{Best (minimum) training loss achieved by transformer and layered reservoir (ReservoirFFN) language models as a function of the number of trainable parameters (log scale). Straight lines show least-squares fits of the form $\mathcal{L}_{\min} = a \log_{10} N_{\mathrm{param}} + b$, revealing power-law-like scaling for both architectures, with a steeper slope for transformers.}
    \label{fig:layered_scaling}
\end{figure*}

\paragraph*{Fixed reservoir.}
Within each block, the incoming representation is first normalized and then passed through a single-layer recurrent reservoir with a tanh nonlinearity. The reservoir parameters are randomly initialized once and remain fixed throughout training. Following standard echo-state principles, the recurrent dynamics are scaled to ensure stable temporal behavior, with the effective memory depth increasing mildly with layer index to provide a hierarchy of time scales across depth.

The reservoir state is normalized before further processing. No gradients are propagated through the reservoir itself; instead, learning occurs exclusively in the surrounding trainable components.

\paragraph*{Reservoir MLP and scaling.}
To map the reservoir state back to the model dimension, each block contains a small trainable MLP. This MLP is intentionally narrow, serving primarily as a readout that extracts useful features from the high-dimensional reservoir dynamics. A learned scalar controls the overall contribution of the reservoir path, allowing the model to adaptively adjust the influence of reservoir-induced features during training. This mechanism stabilizes optimization by preventing overly strong reservoir effects early in training.

\paragraph*{Input-dependent mixing gate.}
The block combines the original representation and the reservoir-refined representation using an input-dependent gating mechanism. The gate operates at the feature level and determines, for each token, how strongly the model should rely on reservoir-induced updates versus the original signal given by a residual connection. This allows the architecture to interpolate smoothly between purely feed-forward behavior and strong temporal mixing, depending on context.

\paragraph*{Feed-forward refinement}
Following the mixing step, each block applies a standard position-wise feed-forward network as is typical for a transformer setup. The feed forward dimension is set to 4 times the embedding dimensions.

\paragraph*{Output layer.}
After passing through all $L$ blocks, the final representation is normalized and projected back into vocabulary space to produce next-token logits. The model is trained using the standard next-token cross-entropy objective over BPE-tokenized text, identical to conventional transformer language models.

Overall, the resulting \emph{Reservoir Feed-Forward Network (FFN)} replaces transformer self-attention with fixed recurrent reservoirs, while preserving the depth, normalization structure, and feed-forward refinement characteristic of modern LLM architectures. All trainable parameters reside in the embeddings, gating networks, small MLPs, feed-forward layers, and output projection.

\subsection{Training setup on large-scale token data}

To evaluate the layered reservoir architecture in a realistic LLM setting, we train it on a byte-pair encoded corpus derived from OpenWebText using a GPT-2-style vocabulary and a fixed context length. The data are stored as memory-mapped token streams for efficient sampling of contiguous training sequences.

We consider a family of models in which the embedding dimension, reservoir size, and number of layers are scaled jointly to produce parameter counts comparable to compact transformer baselines. All models are trained using the same hyperparameters. See Appendix \ref{sec:app_1} for details.

\subsection{Results: optimization behavior and scaling laws}

Figure~\ref{fig:layered_loss} shows the training loss as a function of optimization step for both transformer baselines and LAERC models across multiple model sizes. As expected, transformers achieve lower absolute loss at a fixed parameter budget. However, the layered reservoir models exhibit stable and well-behaved optimization dynamics across all evaluated configurations, with performance consistently improving as model capacity increases.

To quantify how the layered reservoir models scale with size and compare with the transformer scaling \cite{kaplan2020scaling}, we extract, for each configuration, the best (minimum) training loss attained during optimization and plot it against the number of trainable parameters on a logarithmic scale, as shown in Fig.~\ref{fig:layered_scaling}. Both transformers and LAERC models obey approximate power-law scaling relationships of the form
\begin{equation}
    \mathcal{L}_{\min}(N_{\mathrm{param}}) \approx a \log_{10} N_{\mathrm{param}} + b,
\end{equation}
where $N_{\mathrm{param}}$ denotes the total number of trainable parameters and $(a,b)$ are fitting coefficients. For the transformer baselines we obtain a slope $a_{\mathrm{TF}} \approx -1.72$, which is a high scaling law value raising from the smaller parameter regime investigated here, while the LAERC models exhibit a weaker but still substantial slope $a_{\mathrm{Res}} \approx -0.81$.

While these results indicate systematic scaling behavior, the present LAERC implementation should be viewed as a baseline rather than an optimized architecture. Several design choices (e.g., reservoir parameterization, coupling topology, and training protocol) were not exhaustively tuned in this study. Future work can therefore focus on architectural and optimization refinements to assess the extent to which the observed scaling trends can be strengthened.

\section{Conclusion}
\label{sec:conclusion}
We presented a unified framework for comparing classical reservoir computing (with a static linear readout), attention-enhanced reservoir architectures (with a dynamic readout), and transformer architectures on a small character-level Shakespeare corpus for edge computing with low computation. By systematically varying model sizes and evaluating cross-entropy, N-gram overlap, and training/inference time cost, we demonstrated how each architecture scales in performance and computational efficiency. Transformers excel overall, confirming their state-of-the-art status, although this performance comes at a higher computational cost due to the quadratic complexity of attention with respect to the sequence length. Traditional reservoirs, while inherently limited in their ability to capture complex long-range dependencies, nevertheless exhibit unexpectedly competitive results considering their simplicity and low computational overhead. Attention-enhanced reservoirs strike an intermediate balance, significantly improving over classical reservoirs while approaching transformer-level performance with adjustable complexity that depends linearly on sequence length.

We additionally extended the analysis to a large-scale token-level language modeling setting using the proposed \emph{Layered Attention-Enhanced Reservoir Computer} (LAERC). By interleaving multiple fixed recurrent reservoirs with lightweight gating and feed-forward refinement, the layered architecture substantially improves over the single-reservoir setup enabling scaling to larger models. In particular, empirical scaling-law analysis confirms that the layered reservoir models follow a power-law trend in performance as a function of the number of trainable parameters. Though numerically weaker compared to the transformer, the LAERC enables the temporal mixing layer to be implemented with any physical substrate as reservoir, which would substitute the attention layer in a classical transformer.

\section{Data Availability}

The code used for the Shakespeare simulations is available at \url{https://github.com/fekoester/Shakespeare_Res}
and the implementation of the layered reservoir language model for token-level LLM training is provided at \url{https://github.com/fekoester/LAERC}.

\begin{acknowledgments}
This study was supported in part by JSPS KAKENHI (JP22H05195,
JP24KF0179, JP25H01129) and JST CREST (JPMJCR24R2) in Japan.
\end{acknowledgments}

\bibliography{bib}

\appendix


\section{Appendix: LAERC technical specification and training details}
\label{sec:app_1}

This appendix provides a self-contained technical definition of the layered reservoir language model (LAERC) used in Sec.~\ref{sec:layered_reservoir_lm} and summarizes the exact training and implementation settings used for the OpenWebText simulations.

\subsection{Model definition}
\label{sec:app_laerc_model}

Let $V$ denote the vocabulary size, $D$ the embedding dimension, $L$ the number of stacked LAERC blocks, and $R$ the reservoir dimension. We consider a token sequence $X=(x_1,\dots,x_T)$ with $x_t\in\{1,\dots,V\}$.

Tokens are embedded via an embedding matrix $\mathbf{E}\in\mathbb{R}^{D\times V}$,
\begin{equation}
\mathbf{x}_t^{(1)}=\mathbf{E}\,\mathbf{e}(x_t)\in\mathbb{R}^D,
\end{equation}
where $\mathbf{e}(x_t)$ is the one-hot encoding. We use weight tying, i.e., the output projection shares parameters with the input embedding: $\mathbf{W}_{\mathrm{out}}=\mathbf{E}^\top$.

For layer $\ell\in\{1,\dots,L\}$ and time step $t$, the block input is $\mathbf{x}_t^{(\ell)}\in\mathbb{R}^D$. Each block combines (i) a fixed recurrent reservoir path, (ii) an input-dependent mixing gate, and (iii) a feed-forward refinement, consistent with Fig.~\ref{fig:laerc_arch}(b).

\subsubsection*{(i) Fixed reservoir path}
The incoming representation is first normalized:
\begin{equation}
\tilde{\mathbf{x}}_t^{(\ell)}=\mathrm{LN}_{\mathrm{res}}^{(\ell)}(\mathbf{x}_t^{(\ell)})\in\mathbb{R}^D.
\end{equation}
A single-layer tanh RNN reservoir (hidden size $R$) evolves as
\begin{equation}
\mathbf{s}_t^{(\ell)}=\tanh\!\Big(\mathbf{W}_{\mathrm{res}}^{(\ell)}\mathbf{s}_{t-1}^{(\ell)}+\mathbf{W}_{\mathrm{in}}^{(\ell)}\tilde{\mathbf{x}}_t^{(\ell)}+\mathbf{b}^{(\ell)}\Big)\in\mathbb{R}^R,
\end{equation}
where $\mathbf{W}_{\mathrm{res}}^{(\ell)}\in\mathbb{R}^{R\times R}$, $\mathbf{W}_{\mathrm{in}}^{(\ell)}\in\mathbb{R}^{R\times D}$, and $\mathbf{b}^{(\ell)}\in\mathbb{R}^R$.

All reservoir parameters are initialized once and then fixed (no training, and gradients are not propagated through the reservoir dynamics). The recurrent matrix is scaled so that its spectral radius satisfies
\begin{equation}
\rho(\mathbf{W}_{\mathrm{res}}^{(\ell)})\approx \rho_\ell<1.
\end{equation}
In our simulations, we use a linear depth schedule
\begin{equation}
\rho_\ell = 0.95 + 0.04\cdot\frac{\ell-1}{\max(1,L-1)} \qquad (\ell=1,\dots,L),
\end{equation}
which slightly increases memory with depth.

The reservoir state is normalized prior to readout:
\begin{equation}
\tilde{\mathbf{s}}_t^{(\ell)}=\mathrm{LN}_{\mathrm{res,out}}^{(\ell)}(\mathbf{s}_t^{(\ell)})\in\mathbb{R}^R.
\end{equation}

To map from reservoir space back to model dimension, each block contains a narrow MLP:
\begin{align}
\mathbf{u}_t^{(\ell)} &= \phi\!\left(\mathbf{W}^{(\ell)}_{1,\mathrm{res}}\tilde{\mathbf{s}}_t^{(\ell)}+\mathbf{b}^{(\ell)}_{1,\mathrm{res}}\right)\in\mathbb{R}^{H_{\mathrm{res}}},\\
\mathbf{r}_t^{(\ell)} &= \exp(\beta^{(\ell)})\left(\mathbf{W}^{(\ell)}_{2,\mathrm{res}}\mathbf{u}_t^{(\ell)}+\mathbf{b}^{(\ell)}_{2,\mathrm{res}}\right)\in\mathbb{R}^{D},
\end{align}
where $\phi$ is GELU, $H_{\mathrm{res}}=\alpha_{\mathrm{res}}R$ is the hidden width multiplier, and $\beta^{(\ell)}$ is a learned scalar controlling the reservoir-path strength. We initialize $\beta^{(\ell)}\approx 0$ so that the reservoir readout enters at unit scale while remaining trainable.

\subsubsection*{(ii) Input-dependent mixing gate}
The block combines the original representation $\mathbf{x}_t^{(\ell)}$ and reservoir-refined representation $\mathbf{r}_t^{(\ell)}$ via a feature-wise sigmoid gate:
\begin{align}
\mathbf{z}_t^{(\ell)} &= [\mathbf{x}_t^{(\ell)};\mathbf{r}_t^{(\ell)}]\in\mathbb{R}^{2D},\\
\hat{\mathbf{z}}_t^{(\ell)} &= \mathrm{LN}_{\mathrm{mix}}^{(\ell)}(\mathbf{z}_t^{(\ell)}),\\
\mathbf{g}_t^{(\ell)} &= \sigma\!\left(\mathbf{W}_g^{(\ell)}\hat{\mathbf{z}}_t^{(\ell)}+\mathbf{b}_g^{(\ell)}\right)\in\mathbb{R}^{D},\\
\mathbf{m}_t^{(\ell)} &= \mathbf{g}_t^{(\ell)}\odot \mathbf{r}_t^{(\ell)} + \left(\mathbf{1}-\mathbf{g}_t^{(\ell)}\right)\odot \mathbf{x}_t^{(\ell)}\in\mathbb{R}^{D}.
\end{align}

\subsubsection*{(iii) Feed-forward refinement}
A standard position-wise feed-forward network refines the mixed representation:
\begin{align}
&\hat{\mathbf{m}}_t^{(\ell)} = \mathrm{LN}_{\mathrm{ffn}}^{(\ell)}(\mathbf{m}_t^{(\ell)}),\\
&\mathbf{h}_t^{(\ell)} = \phi\!\left(\mathbf{W}^{(\ell)}_{1,\mathrm{ffn}}\hat{\mathbf{m}}_t^{(\ell)}+\mathbf{b}^{(\ell)}_{1,\mathrm{ffn}}\right)\in\mathbb{R}^{H_{\mathrm{ffn}}},\\
&\mathbf{f}_t^{(\ell)} = \mathbf{W}^{(\ell)}_{2,\mathrm{ffn}}\mathbf{h}_t^{(\ell)}+\mathbf{b}^{(\ell)}_{2,\mathrm{ffn}}\in\mathbb{R}^{D},\\
&\mathbf{x}_t^{(\ell+1)} = \mathbf{m}_t^{(\ell)} + \alpha^{(\ell)}\mathbf{f}_t^{(\ell)}.
\end{align}
We initialize $\alpha^{(\ell)}=0$, making each block close to the identity map at initialization and improving stability for deep stacks.

After $L$ blocks, a final LayerNorm and tied projection produce logits:
\begin{align}
&\mathbf{z}_t = \mathrm{LN}_f(\mathbf{x}_t^{(L+1)}),\\
&\boldsymbol{\ell}_t = \mathbf{E}^\top \mathbf{z}_t\in\mathbb{R}^V,\\
&p(y_t=v\mid X)=\mathrm{softmax}(\boldsymbol{\ell}_t)_v.
\end{align}
Training minimizes standard next-token cross-entropy over the tokenized corpus.

\subsection{Training data and batching}
\label{sec:app_laerc_data}

We train on an OpenWebText-derived corpus tokenized with a GPT-2-style BPE vocabulary ($V=50{,}257$) and fixed context length $T=512$. The dataset is stored as contiguous memory-mapped integer token streams. Each microbatch is generated by sampling $B$ random start indices uniformly from the training stream, extracting contiguous chunks of length $T$, and forming next-token prediction pairs:
\begin{equation}
\mathbf{x} = (x_1,\dots,x_{T-1}),\qquad \mathbf{y}=(x_2,\dots,x_T).
\end{equation}

\subsection{Hyperparameters and implementation details}
\label{sec:app_laerc_hparams}

Unless otherwise stated, we use $R=4D$ and a linear spectral-radius schedule $\rho_\ell\in[0.95,0.99]$ across depth. The reservoir readout MLP width is set to $H_{\mathrm{res}}=\alpha_{\mathrm{res}}R$ with $\alpha_{\mathrm{res}}=1.0$.

We train with AdamW using $(\beta_1,\beta_2)=(0.9,0.999)$. We apply weight decay ($0.1$) to linear weights and use no weight decay for biases, LayerNorm parameters, and embedding parameters.

We report the effective global batch size as
\begin{equation}
B_{\mathrm{eff}} = B \times \mathrm{accum},
\end{equation}
where $B=32$ is the microbatch size and $\mathrm{accum}=4$ is the number of gradient accumulation steps. For the used context length of $L=512$ thus 65536 tokens per gradient steps are used.

Training uses mixed precision on GPUs (bfloat16 when supported, otherwise float16 with dynamic loss scaling). We enable TF32 matrix multiplications on Ampere-class GPUs and cuDNN benchmarking for kernel selection. Reservoir parameters are stored in low precision when possible and remain fixed throughout training.

Embedding and linear weights are initialized from a normal distribution with standard deviation $0.02$. Output projection is tied to the embedding matrix.

\end{document}